\documentclass[10pt, journal, twocolumn]{IEEEtran}

\usepackage{color,graphicx,amsmath,amssymb,amsthm,epsfig,mathrsfs,cite,bm,graphics}
\usepackage{caption}
\usepackage{algpseudocode}
\usepackage{algorithmicx}
\usepackage{algorithm} 
\usepackage[utf8]{inputenc}
\usepackage{amsmath}
\usepackage{amsfonts}
\usepackage{amssymb}
\usepackage{placeins}
\usepackage{graphicx}
\usepackage{bbm}
\usepackage{graphicx,graphics,subcaption}
\usepackage{url}
\usepackage{tikz}
\usetikzlibrary{automata, positioning, arrows,shapes.multipart}
\usepackage{amsmath,amsthm,bm}
\usepackage{amssymb}
\newcommand{\norm}[1]{\left\lVert#1\right\rVert}
\usepackage{xcolor}
\usepackage{graphics,graphicx,color,epsfig}
\usepackage{tikz}
\usetikzlibrary{automata, positioning, arrows,shapes.multipart,fit}
\usepackage{pgfplots}
\usepackage{lipsum}
\usepackage{mathtools}
\usepackage{cuted}
\usepackage{braket}
\usepackage{booktabs}
\DeclareMathOperator{\tr}{Tr}

\newcommand{\cutt}[1]{}
\newcommand{\cut}[1]{}
\usepackage{multirow}

\title{Deep learned SVT: Unrolling singular value thresholding to obtain better MSE}
\author{ Siva Shanmugam* \thanks{*The authors are with the Department of Electrical Engineering, Indian Institute of Technology Madras, Chennai, India 600 036. Email: \texttt{ee17s024@smail.iitm.ac.in,skalyani@ee.iitm.ac.in}} \hspace{16pt} Sheetal Kalyani*}
\pgfplotsset{compat=1.17}
\begin{document}
	\maketitle
    \begin{abstract}
    Affine rank minimization problem is the generalized version of low rank matrix completion problem where linear combinations of the entries of a low rank matrix are observed and the matrix is estimated from these measurements. We propose a trainable deep neural network by unrolling a popular iterative algorithm called the singular value thresholding (SVT) algorithm to perform this generalized matrix completion which we call Learned SVT (LSVT). We show that our proposed LSVT with fixed layers (say $T$) reconstructs the matrix with lesser mean squared error (MSE) compared with that incurred by SVT with fixed (same $T$) number of iterations and our method is much more robust to the parameters which need to be carefully chosen in SVT algorithm. 
\end{abstract}
\section{Introduction}
The problem of completing a low rank matrix by sampling only a few of its entries is a well studied problem which finds its application in variety of areas a few of which include Euclidean distance matrix completion \cite{localizationedm}, environmental monitoring using sensors \cite{environmental} \cite{lowcostwsnmc}, array signal processing \cite{arraysignal}, beamforming \cite{beamforming} and wireless channel estimation \cite{mimomc}. In \cite{exact} the authors showed that its possible to perfectly complete low rank matrix by observing only a few entries and by solving a convex optimization problem. 

The generalized low rank matrix completion problem in which a number of linear combinations of its entries are measured as opposed to sampling the entries directly has also attracted a lot of attention. The problem of reconstructing finite dimensional quantum states is naturally a generalized matrix completion problem \cite{gross}, \cite{sevenqubit}. In \cite{arm}, the affine rank minimization problem which minimizes the rank of the matrix with affine constraints was proposed for solving this generalized matrix completion. The work in \cite{lowrankanybasisgross} quantified the number of measurements required for the success of matrix recovery from its linear measurements. In \cite{powerfactorization} authors proposed the power factorization algorithm for generalized matrix completion and the idea was further extended in \cite{als}. In \cite{saresepowerfactorization}, the authors extended the power factorization algorithm to estimate sparse low rank matrix whose left and right singular matrices are sparse. The recovery of positive semidefinite matrix in the context of generalized matrix completion has been analyzed in \cite{rankonemeas} and \cite{psdrlm}. The singular value thresholding (SVT) \cite{svt} algorithm is a very popular algorithm used for performing matrix completion and generalized matrix completion.

The capability of computing machines to store and process a huge amount of data and the capability of neural networks to learn very complicated functions have enabled the deep networks to find its roots in almost all fields. However, it has been very hard to interpret these deep neural networks. In this context, unrolled algorithms which are deep neural networks whose architecture is inspired from interpretable classical algorithms have attracted considerable attention recently \cite{unrollingsurvey}. We present a deep neural network inspired from SVT algorithm for general matrix completion which reconstructs matrix with significantly lower MSE and is more robust to the parameters that need to be carefully chosen in SVT.

\subsection{Our contributions and the outline of work}
In this paper, we design a trainable deep neural network to perform affine rank minimization by unrolling the SVT algorithm. Each layer of the deep network is similar to a single iteration of the SVT algorithm except that the parameters such as measurement matrices, the step sizes, threshold values used in the SVT are now learnable. We term the proposed method Learned SVT (LSVT). The advantages of the proposed method are as follows. Firstly, the proposed LSVT outperforms SVT meaning that our network with $T$ layers reconstructs the matrix with lesser mean squared error (MSE) compared with the MSE incurred by SVT with fixed (same $T$) number of iterations. Secondly, LSVT seems more robust to the initialization than SVT in all our empirical results.

\vspace{-3mm}
\subsection{Notations used}
Matrices, vectors and scalars are represented by upper case, bold lower case and lower case respectively. $\tr[A], \norm{A}_F$ and $\norm{A}_{tr}$ denote trace, Frobenius norm and nuclear norm of matrix $A$ respectively where nuclear norm is the sum of singular values of $A$. The $i^{th}$ element of $\bm b$ is denoted by $b_i$ and $\norm{\bm b}_2$ denotes the Euclidean norm of $\bm b$.

    \section{Problem formulation} \label{sec:problem}
Let $X \in \mathbb{R}^{d \times d}$ be the true matrix to be recovered and $r$ be the rank of $X$. Note that, in general, we need $d^2$ measurements of the matrix $X$ to get the complete information about the matrix $X$. This holds when the matrix is of full rank. But the low rank structure enables the recovery of $X$ from its fewer than $d^2$ measurements. For a rank-$r$ matrix, the degree of freedom reduces from $d^2$ to $r \times (2d-r)$. Let $\{A_i \in \mathbb{R}^{d \times d}\}_{i=1}^{m}$  be a set of $m$ ($m<d^2$) measurement matrices such that $\tr[A_i^T A_j] = \delta_{ij}$. Let $\bm b \in \mathbb{R}^m$ be $m$ linear measurements of the matrix $X$ which are given as
\begin{equation}
    b_i = \tr[A_i X], \quad \forall \, i = 1, \dots m \label{eqn:meas}
\end{equation}
In other words, $\bm b$ is the linear function of the unknown matrix $X$ where the linear map $\mathcal{A}: \mathbb{R}^{d \times d} \to \mathbb{R}^m$ (also called the sampling operator) is defined as
\begin{equation}
    \mathcal{A}(X) = \begin{bmatrix} \tr[A_1 X] & \tr[A_2 X] & \dots & \tr[A_m X] \end{bmatrix}^T \label{eqn:lmap}
\end{equation}
The problem that we consider in this paper is to recover the matrix $X$ from the measurements $\bm b$ assuming that the measurement matrices are known. The matrix recovery is done by solving the affine rank minimization problem given as
\begin{subequations}
    \begin{alignat}{2}
    &\!\min_{X \in \mathbb{R}^{d \times d}}        &\qquad& \text{Rank}(X)  \\
    &\text{such that} &      & \mathcal{A}(X)=\bm b
    \end{alignat}  \label{eqn:arm}
\end{subequations}
where $\mathcal{A}$ is the linear map as defined in \eqref{eqn:lmap}. The affine rank minimization problem \eqref{eqn:arm} minimizes the rank of the matrix within an affine constraint set $\{X: \mathcal{A}(X)=\bm b\}$ which is a level set of the linear map $\mathcal{A}$.

We design a deep neural network inspired from SVT \cite{svt} algorithm to solve the affine rank minimization formulated in \eqref{eqn:arm}.  First, we present the SVT algorithm in the next section then in subsequent sections we design a neural network by unrolling the iterations of the SVT algorithm. 

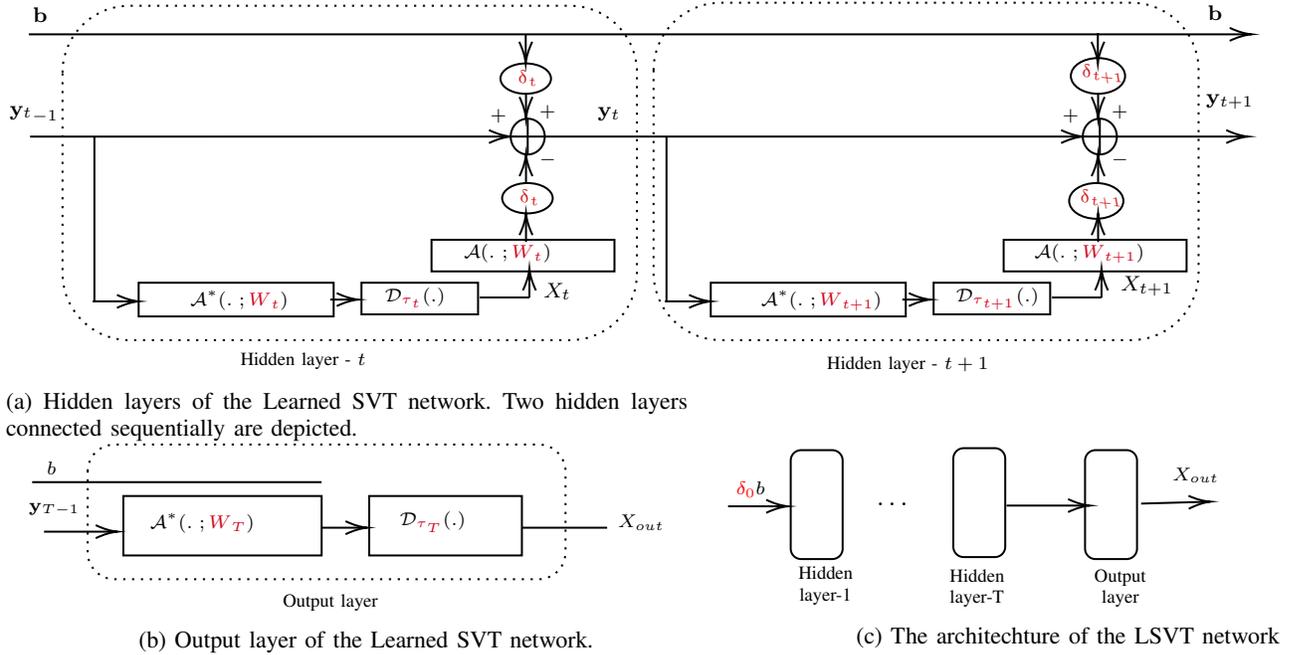
\begin{figure*}[ht!]
\hspace{9mm}
\begin{subfigure}[H]{.5\textwidth}
\centering
\tikzset{every picture/.style={line width=0.75pt}} 

\begin{tikzpicture}[x=0.75pt,y=0.75pt,yscale=-1,xscale=1]

\draw    (98.34,161.59) -- (119.51,161.59) ;
\draw [shift={(121.51,161.59)}, rotate = 180] [color={rgb, 255:red, 0; green, 0; blue, 0 }  ][line width=0.75]    (10.93,-3.29) .. controls (6.95,-1.4) and (3.31,-0.3) .. (0,0) .. controls (3.31,0.3) and (6.95,1.4) .. (10.93,3.29)   ;
\draw    (318.59,159.63) -- (318.59,147.12) ;
\draw [shift={(318.59,145.12)}, rotate = 450] [color={rgb, 255:red, 0; green, 0; blue, 0 }  ][line width=0.75]    (10.93,-3.29) .. controls (6.95,-1.4) and (3.31,-0.3) .. (0,0) .. controls (3.31,0.3) and (6.95,1.4) .. (10.93,3.29)   ;
\draw   (269.23,130.51) -- (361.49,130.51) -- (361.49,146.46) -- (269.23,146.46) -- cycle ;
\draw    (99.2,78.44) -- (99.2,161.59) ;
\draw   (309.14,78.41) .. controls (309.14,73.42) and (312.99,69.37) .. (317.73,69.37) .. controls (322.47,69.37) and (326.32,73.42) .. (326.32,78.41) .. controls (326.32,83.4) and (322.47,87.45) .. (317.73,87.45) .. controls (312.99,87.45) and (309.14,83.4) .. (309.14,78.41) -- cycle ; \draw   (309.14,78.41) -- (326.32,78.41) ; \draw   (317.73,69.37) -- (317.73,87.45) ;
\draw    (316.73,27.08) -- (316.73,39.5) ;
\draw [shift={(316.73,41.5)}, rotate = 270] [color={rgb, 255:red, 0; green, 0; blue, 0 }  ][line width=0.75]    (10.93,-3.29) .. controls (6.95,-1.4) and (3.31,-0.3) .. (0,0) .. controls (3.31,0.3) and (6.95,1.4) .. (10.93,3.29)   ;
\draw    (316.87,131.6) -- (316.87,119.18) ;
\draw [shift={(316.87,117.18)}, rotate = 450] [color={rgb, 255:red, 0; green, 0; blue, 0 }  ][line width=0.75]    (10.93,-3.29) .. controls (6.95,-1.4) and (3.31,-0.3) .. (0,0) .. controls (3.31,0.3) and (6.95,1.4) .. (10.93,3.29)   ;
\draw    (293.56,159.63) -- (318.59,159.63) ;
\draw  [dash pattern={on 0.84pt off 2.51pt}] (83.13,46.15) .. controls (83.13,27.51) and (98.24,12.39) .. (116.89,12.39) -- (339.08,12.39) .. controls (357.72,12.39) and (372.83,27.51) .. (372.83,46.15) -- (372.83,147.41) .. controls (372.83,166.05) and (357.72,181.16) .. (339.08,181.16) -- (116.89,181.16) .. controls (98.24,181.16) and (83.13,166.05) .. (83.13,147.41) -- cycle ;
\draw   (233.89,152.18) -- (292.7,152.18) -- (292.7,168.14) -- (233.89,168.14) -- cycle ;
\draw   (121.51,152.18) -- (219.86,152.18) -- (219.86,169.19) -- (121.51,169.19) -- cycle ;
\draw    (218.14,160.69) -- (230.33,160.69) ;
\draw [shift={(232.33,160.69)}, rotate = 180] [color={rgb, 255:red, 0; green, 0; blue, 0 }  ][line width=0.75]    (10.93,-3.29) .. controls (6.95,-1.4) and (3.31,-0.3) .. (0,0) .. controls (3.31,0.3) and (6.95,1.4) .. (10.93,3.29)   ;
\draw  [dash pattern={on 0.84pt off 2.51pt}] (381.22,44.92) .. controls (381.22,26.11) and (396.47,10.86) .. (415.28,10.86) -- (622.02,10.86) .. controls (640.83,10.86) and (656.08,26.11) .. (656.08,44.92) -- (656.08,147.1) .. controls (656.08,165.91) and (640.83,181.16) .. (622.02,181.16) -- (415.28,181.16) .. controls (396.47,181.16) and (381.22,165.91) .. (381.22,147.1) -- cycle ;
\draw    (66.75,78.43) -- (307.14,78.41) ;
\draw [shift={(309.14,78.41)}, rotate = 540] [color={rgb, 255:red, 0; green, 0; blue, 0 }  ][line width=0.75]    (10.93,-3.29) .. controls (6.95,-1.4) and (3.31,-0.3) .. (0,0) .. controls (3.31,0.3) and (6.95,1.4) .. (10.93,3.29)   ;
\draw    (66.11,26.81) -- (680.7,26.81) ;
\draw [shift={(682.7,26.81)}, rotate = 180] [color={rgb, 255:red, 0; green, 0; blue, 0 }  ][line width=0.75]    (10.93,-3.29) .. controls (6.95,-1.4) and (3.31,-0.3) .. (0,0) .. controls (3.31,0.3) and (6.95,1.4) .. (10.93,3.29)   ;
\draw   (304.85,109.53) .. controls (304.85,105.3) and (310.55,101.87) .. (317.59,101.87) .. controls (324.63,101.87) and (330.34,105.3) .. (330.34,109.53) .. controls (330.34,113.76) and (324.63,117.18) .. (317.59,117.18) .. controls (310.55,117.18) and (304.85,113.76) .. (304.85,109.53) -- cycle ;
\draw    (316.87,101.87) -- (316.87,89.45) ;
\draw [shift={(316.87,87.45)}, rotate = 450] [color={rgb, 255:red, 0; green, 0; blue, 0 }  ][line width=0.75]    (10.93,-3.29) .. controls (6.95,-1.4) and (3.31,-0.3) .. (0,0) .. controls (3.31,0.3) and (6.95,1.4) .. (10.93,3.29)   ;
\draw   (303.99,49.16) .. controls (303.99,44.93) and (309.7,41.5) .. (316.73,41.5) .. controls (323.77,41.5) and (329.48,44.93) .. (329.48,49.16) .. controls (329.48,53.39) and (323.77,56.81) .. (316.73,56.81) .. controls (309.7,56.81) and (303.99,53.39) .. (303.99,49.16) -- cycle ;
\draw    (316.87,56.81) -- (316.87,67.37) ;
\draw [shift={(316.87,69.37)}, rotate = 270] [color={rgb, 255:red, 0; green, 0; blue, 0 }  ][line width=0.75]    (10.93,-3.29) .. controls (6.95,-1.4) and (3.31,-0.3) .. (0,0) .. controls (3.31,0.3) and (6.95,1.4) .. (10.93,3.29)   ;
\draw    (386.88,161.59) -- (408.05,161.59) ;
\draw [shift={(410.05,161.59)}, rotate = 180] [color={rgb, 255:red, 0; green, 0; blue, 0 }  ][line width=0.75]    (10.93,-3.29) .. controls (6.95,-1.4) and (3.31,-0.3) .. (0,0) .. controls (3.31,0.3) and (6.95,1.4) .. (10.93,3.29)   ;
\draw    (607.13,159.63) -- (607.13,147.12) ;
\draw [shift={(607.13,145.12)}, rotate = 450] [color={rgb, 255:red, 0; green, 0; blue, 0 }  ][line width=0.75]    (10.93,-3.29) .. controls (6.95,-1.4) and (3.31,-0.3) .. (0,0) .. controls (3.31,0.3) and (6.95,1.4) .. (10.93,3.29)   ;
\draw   (557.77,130.51) -- (650.03,130.51) -- (650.03,146.46) -- (557.77,146.46) -- cycle ;
\draw    (387.74,78.44) -- (387.74,161.59) ;
\draw   (597.68,78.41) .. controls (597.68,73.42) and (601.53,69.37) .. (606.27,69.37) .. controls (611.01,69.37) and (614.86,73.42) .. (614.86,78.41) .. controls (614.86,83.4) and (611.01,87.45) .. (606.27,87.45) .. controls (601.53,87.45) and (597.68,83.4) .. (597.68,78.41) -- cycle ; \draw   (597.68,78.41) -- (614.86,78.41) ; \draw   (606.27,69.37) -- (606.27,87.45) ;
\draw    (605.28,26.18) -- (605.28,39.5) ;
\draw [shift={(605.28,41.5)}, rotate = 270] [color={rgb, 255:red, 0; green, 0; blue, 0 }  ][line width=0.75]    (10.93,-3.29) .. controls (6.95,-1.4) and (3.31,-0.3) .. (0,0) .. controls (3.31,0.3) and (6.95,1.4) .. (10.93,3.29)   ;
\draw    (605.41,131.6) -- (605.41,119.18) ;
\draw [shift={(605.41,117.18)}, rotate = 450] [color={rgb, 255:red, 0; green, 0; blue, 0 }  ][line width=0.75]    (10.93,-3.29) .. controls (6.95,-1.4) and (3.31,-0.3) .. (0,0) .. controls (3.31,0.3) and (6.95,1.4) .. (10.93,3.29)   ;
\draw    (582.1,159.63) -- (607.13,159.63) ;
\draw   (522.44,152.18) -- (581.24,152.18) -- (581.24,168.14) -- (522.44,168.14) -- cycle ;
\draw   (410.05,152.18) -- (508.4,152.18) -- (508.4,169.19) -- (410.05,169.19) -- cycle ;
\draw    (506.68,160.69) -- (518.88,160.69) ;
\draw [shift={(520.88,160.69)}, rotate = 180] [color={rgb, 255:red, 0; green, 0; blue, 0 }  ][line width=0.75]    (10.93,-3.29) .. controls (6.95,-1.4) and (3.31,-0.3) .. (0,0) .. controls (3.31,0.3) and (6.95,1.4) .. (10.93,3.29)   ;
\draw    (326.32,78.41) -- (595.68,78.41) ;
\draw [shift={(597.68,78.41)}, rotate = 180] [color={rgb, 255:red, 0; green, 0; blue, 0 }  ][line width=0.75]    (10.93,-3.29) .. controls (6.95,-1.4) and (3.31,-0.3) .. (0,0) .. controls (3.31,0.3) and (6.95,1.4) .. (10.93,3.29)   ;
\draw   (590.81,110.88) .. controls (590.81,105.9) and (596.94,101.87) .. (604.49,101.87) .. controls (612.04,101.87) and (618.16,105.9) .. (618.16,110.88) .. controls (618.16,115.85) and (612.04,119.89) .. (604.49,119.89) .. controls (596.94,119.89) and (590.81,115.85) .. (590.81,110.88) -- cycle ;
\draw    (605.41,101.87) -- (605.41,89.45) ;
\draw [shift={(605.41,87.45)}, rotate = 450] [color={rgb, 255:red, 0; green, 0; blue, 0 }  ][line width=0.75]    (10.93,-3.29) .. controls (6.95,-1.4) and (3.31,-0.3) .. (0,0) .. controls (3.31,0.3) and (6.95,1.4) .. (10.93,3.29)   ;
\draw    (605.41,56.81) -- (605.41,67.37) ;
\draw [shift={(605.41,69.37)}, rotate = 270] [color={rgb, 255:red, 0; green, 0; blue, 0 }  ][line width=0.75]    (10.93,-3.29) .. controls (6.95,-1.4) and (3.31,-0.3) .. (0,0) .. controls (3.31,0.3) and (6.95,1.4) .. (10.93,3.29)   ;
\draw   (591.67,47.8) .. controls (591.67,42.83) and (597.79,38.79) .. (605.34,38.79) .. controls (612.9,38.79) and (619.02,42.83) .. (619.02,47.8) .. controls (619.02,52.78) and (612.9,56.81) .. (605.34,56.81) .. controls (597.79,56.81) and (591.67,52.78) .. (591.67,47.8) -- cycle ;
\draw    (614.86,78.41) -- (678.98,78.41) ;
\draw [shift={(680.98,78.41)}, rotate = 180] [color={rgb, 255:red, 0; green, 0; blue, 0 }  ][line width=0.75]    (10.93,-3.29) .. controls (6.95,-1.4) and (3.31,-0.3) .. (0,0) .. controls (3.31,0.3) and (6.95,1.4) .. (10.93,3.29)   ;

\draw (283.98,131.1) node [anchor=north west][inner sep=0.75pt]  [font=\scriptsize] [align=left] {$\displaystyle \mathcal{A}( .\ ;\textcolor[rgb]{0.82,0.01,0.11}{W}\textcolor[rgb]{0.82,0.01,0.11}{_{t}}) \ $};
\draw (322.18,60.78) node [anchor=north west][inner sep=0.75pt]  [font=\scriptsize] [align=left] {$\displaystyle +$};
\draw (322.18,85.07) node [anchor=north west][inner sep=0.75pt]  [font=\scriptsize] [align=left] {$\displaystyle -$};
\draw (297.48,63.11) node [anchor=north west][inner sep=0.75pt]  [font=\scriptsize] [align=left] {$\displaystyle +$};
\draw (66.72,11.48) node [anchor=north west][inner sep=0.75pt]  [font=\footnotesize] [align=left] {$\displaystyle \mathbf{b}$};
\draw (55.01,61.45) node [anchor=north west][inner sep=0.75pt]  [font=\footnotesize] [align=left] {$\displaystyle {\mathbf{y}_{t}}_{-1}$};
\draw (351.58,62.08) node [anchor=north west][inner sep=0.75pt]  [font=\footnotesize] [align=left] {$\displaystyle {\mathbf{y}_{t}}$};
\draw (324.33,150.78) node [anchor=north west][inner sep=0.75pt]  [font=\footnotesize] [align=left] {$\displaystyle X_{t}$};
\draw (171.38,185.73) node [anchor=north west][inner sep=0.75pt]  [font=\scriptsize] [align=left] {Hidden layer - $\displaystyle t$};
\draw (467.3,187.74) node [anchor=north west][inner sep=0.75pt]  [font=\scriptsize] [align=left] {Hidden layer - $\displaystyle t+1$};
\draw (244.3,152.94) node [anchor=north west][inner sep=0.75pt]  [font=\scriptsize] [align=left] {$\displaystyle \mathcal{D}_{\textcolor[rgb]{0.82,0.01,0.11}{\tau }\textcolor[rgb]{0.82,0.01,0.11}{_{t}}}( .) \ $};
\draw (268.41,151.16) node [anchor=north west][inner sep=0.75pt]   [align=left] {$ $};
\draw (145.97,154.56) node [anchor=north west][inner sep=0.75pt]  [font=\scriptsize] [align=left] {$\displaystyle \mathcal{A}^{*}( .\ ;\textcolor[rgb]{0.82,0.01,0.11}{W}\textcolor[rgb]{0.82,0.01,0.11}{_{t}}) \ $};
\draw (658.51,55.34) node [anchor=north west][inner sep=0.75pt]  [font=\footnotesize] [align=left] {$\displaystyle {\mathbf{y}_{t+1}}$};
\draw (659.2,10.76) node [anchor=north west][inner sep=0.75pt]  [font=\footnotesize] [align=left] {$\displaystyle \mathbf{b}$};
\draw (311.28,103.14) node [anchor=north west][inner sep=0.75pt]  [font=\scriptsize,rotate=-4.06,xslant=-0.23] [align=left] {$\displaystyle \textcolor[rgb]{0.83,0.07,0.07}{\delta }\textcolor[rgb]{0.83,0.07,0.07}{_{t}}$};
\draw (310.42,42.77) node [anchor=north west][inner sep=0.75pt]  [font=\scriptsize,rotate=-4.06,xslant=-0.23] [align=left] {$\displaystyle \textcolor[rgb]{0.83,0.07,0.07}{\delta }\textcolor[rgb]{0.83,0.07,0.07}{_{t}}$};
\draw (571.81,131.1) node [anchor=north west][inner sep=0.75pt]  [font=\scriptsize] [align=left] {$\displaystyle \mathcal{A}( .\ ;\textcolor[rgb]{0.82,0.01,0.11}{W}\textcolor[rgb]{0.82,0.01,0.11}{_{t+1}}) \ $};
\draw (610.72,60.78) node [anchor=north west][inner sep=0.75pt]  [font=\scriptsize] [align=left] {$\displaystyle +$};
\draw (610.72,85.07) node [anchor=north west][inner sep=0.75pt]  [font=\scriptsize] [align=left] {$\displaystyle -$};
\draw (586.02,63.11) node [anchor=north west][inner sep=0.75pt]  [font=\scriptsize] [align=left] {$\displaystyle +$};
\draw (615.53,147.18) node [anchor=north west][inner sep=0.75pt]  [font=\footnotesize] [align=left] {$\displaystyle X_{t+1}$};
\draw (532.35,152.94) node [anchor=north west][inner sep=0.75pt]  [font=\scriptsize] [align=left] {$\displaystyle \mathcal{D}_{\textcolor[rgb]{0.82,0.01,0.11}{\tau }\textcolor[rgb]{0.82,0.01,0.11}{_{t+1}}}( .) \ $};
\draw (556.95,151.16) node [anchor=north west][inner sep=0.75pt]   [align=left] {$ $};
\draw (433.88,154.56) node [anchor=north west][inner sep=0.75pt]  [font=\scriptsize] [align=left] {$\displaystyle \mathcal{A}^{*}( .\ ;\textcolor[rgb]{0.82,0.01,0.11}{W}\textcolor[rgb]{0.82,0.01,0.11}{_{t+1}}) \ $};
\draw (593.93,103.17) node [anchor=north west][inner sep=0.75pt]  [font=\scriptsize,rotate=-3.74,xslant=-0.23] [align=left] {$\displaystyle \textcolor[rgb]{0.83,0.07,0.07}{\delta }\textcolor[rgb]{0.83,0.07,0.07}{_{t+1}}$};
\draw (594.79,40.1) node [anchor=north west][inner sep=0.75pt]  [font=\scriptsize,rotate=-3.74,xslant=-0.23] [align=left] {$\displaystyle \textcolor[rgb]{0.83,0.07,0.07}{\delta }\textcolor[rgb]{0.83,0.07,0.07}{_{t+1}}$};
\end{tikzpicture}

\caption{Hidden layers of the Learned SVT network. Two hidden layers connected sequentially are depicted.}
\label{fig:hiddenlayers}
\end{subfigure} \\

\hspace{12mm}
\begin{subfigure}[H]{.5\textwidth}

\tikzset{every picture/.style={line width=0.75pt}} 

\begin{tikzpicture}[x=0.75pt,y=0.75pt,yscale=-1,xscale=1]

\draw   (334,101) -- (411,101) -- (411,131) -- (334,131) -- cycle ;
\draw   (210,101) -- (310,101) -- (310,131) -- (210,131) -- cycle ;
\draw    (170,119) -- (205,119) ;
\draw [shift={(207,119)}, rotate = 180] [color={rgb, 255:red, 0; green, 0; blue, 0 }  ][line width=0.75]    (10.93,-3.29) .. controls (6.95,-1.4) and (3.31,-0.3) .. (0,0) .. controls (3.31,0.3) and (6.95,1.4) .. (10.93,3.29)   ;
\draw    (310,117) -- (331,117) ;
\draw [shift={(333,117)}, rotate = 180] [color={rgb, 255:red, 0; green, 0; blue, 0 }  ][line width=0.75]    (10.93,-3.29) .. controls (6.95,-1.4) and (3.31,-0.3) .. (0,0) .. controls (3.31,0.3) and (6.95,1.4) .. (10.93,3.29)   ;
\draw    (411,117) -- (454,117) ;
\draw  [dash pattern={on 0.84pt off 2.51pt}] (192,88.6) .. controls (192,81.09) and (198.09,75) .. (205.6,75) -- (419.4,75) .. controls (426.91,75) and (433,81.09) .. (433,88.6) -- (433,129.4) .. controls (433,136.91) and (426.91,143) .. (419.4,143) -- (205.6,143) .. controls (198.09,143) and (192,136.91) .. (192,129.4) -- cycle ;
\draw    (164,94) -- (310,94) ;

\draw (348,107) node [anchor=north west][inner sep=0.75pt]  [font=\scriptsize] [align=left] {$\displaystyle \mathcal{D}_{\textcolor[rgb]{0.82,0.01,0.11}{\tau }\textcolor[rgb]{0.82,0.01,0.11}{_{T}}}( .) \ $};
\draw (367,107) node [anchor=north west][inner sep=0.75pt]   [align=left] {$ $};
\draw (222,108) node [anchor=north west][inner sep=0.75pt]  [font=\scriptsize] [align=left] {$\displaystyle \mathcal{A}^{*}( .\ ;\textcolor[rgb]{0.82,0.01,0.11}{W}\textcolor[rgb]{0.82,0.01,0.11}{_{T}}) \ $};
\draw (161,103) node [anchor=north west][inner sep=0.75pt]  [font=\scriptsize] [align=left] {$\displaystyle \mathbf{y}_{T-1}$};
\draw (458,109) node [anchor=north west][inner sep=0.75pt]  [font=\scriptsize] [align=left] {$\displaystyle X_{out}$};
\draw (289,149) node [anchor=north west][inner sep=0.75pt]  [font=\scriptsize] [align=left] {Output layer};
\draw (170,82) node [anchor=north west][inner sep=0.75pt]  [font=\scriptsize] [align=left] {$\displaystyle b$};

\end{tikzpicture}

    \caption{Output layer of the Learned SVT network.}
    \label{fig:outputlayer}
\end{subfigure}
\hspace{1mm}
\begin{subfigure}[H]{.5\textwidth}
 
\tikzset{every picture/.style={line width=0.75pt}} 

\begin{tikzpicture}[x=0.75pt,y=0.75pt,yscale=-1,xscale=1]

\draw   (102.52,57.24) .. controls (102.52,54.35) and (104.86,52) .. (107.76,52) -- (123.48,52) .. controls (126.37,52) and (128.72,54.35) .. (128.72,57.24) -- (128.72,101.73) .. controls (128.72,104.63) and (126.37,106.97) .. (123.48,106.97) -- (107.76,106.97) .. controls (104.86,106.97) and (102.52,104.63) .. (102.52,101.73) -- cycle ;
\draw    (70.94,80.34) -- (99.1,80.34) ;
\draw [shift={(101.1,80.34)}, rotate = 180] [color={rgb, 255:red, 0; green, 0; blue, 0 }  ][line width=0.75]    (10.93,-3.29) .. controls (6.95,-1.4) and (3.31,-0.3) .. (0,0) .. controls (3.31,0.3) and (6.95,1.4) .. (10.93,3.29)   ;
\draw    (279.25,78.97) -- (312.13,78.06) ;
\draw [shift={(314.13,78)}, rotate = 538.4] [color={rgb, 255:red, 0; green, 0; blue, 0 }  ][line width=0.75]    (10.93,-3.29) .. controls (6.95,-1.4) and (3.31,-0.3) .. (0,0) .. controls (3.31,0.3) and (6.95,1.4) .. (10.93,3.29)   ;
\draw    (211.39,79.97) -- (251.33,79.97) ;
\draw [shift={(253.33,79.97)}, rotate = 180] [color={rgb, 255:red, 0; green, 0; blue, 0 }  ][line width=0.75]    (10.93,-3.29) .. controls (6.95,-1.4) and (3.31,-0.3) .. (0,0) .. controls (3.31,0.3) and (6.95,1.4) .. (10.93,3.29)   ;
\draw   (184.52,56.24) .. controls (184.52,53.35) and (186.87,51) .. (189.76,51) -- (205.48,51) .. controls (208.38,51) and (210.72,53.35) .. (210.72,56.24) -- (210.72,100.73) .. controls (210.72,103.63) and (208.38,105.97) .. (205.48,105.97) -- (189.76,105.97) .. controls (186.87,105.97) and (184.52,103.63) .. (184.52,100.73) -- cycle ;
\draw   (250.97,57.24) .. controls (250.97,54.35) and (253.32,52) .. (256.21,52) -- (271.94,52) .. controls (274.83,52) and (277.18,54.35) .. (277.18,57.24) -- (277.18,101.73) .. controls (277.18,104.63) and (274.83,106.97) .. (271.94,106.97) -- (256.21,106.97) .. controls (253.32,106.97) and (250.97,104.63) .. (250.97,101.73) -- cycle ;

\draw (100.49,109.39) node [anchor=north west][inner sep=0.75pt]  [font=\scriptsize] [align=left] {\begin{minipage}[lt]{27.313356000000002pt}\setlength\topsep{0pt}
\begin{center}
Hidden \\layer-1
\end{center}

\end{minipage}};
\draw (176.83,110.66) node [anchor=north west][inner sep=0.75pt]  [font=\scriptsize] [align=left] {\begin{minipage}[lt]{27.313356000000002pt}\setlength\topsep{0pt}
\begin{center}
Hidden \\layer-T
\end{center}

\end{minipage}};
\draw (83,101.62) node [anchor=north west][inner sep=0.75pt]   [align=left] {$ $};
\draw (73.48,66.43) node [anchor=north west][inner sep=0.75pt]  [font=\scriptsize] [align=left] {$\displaystyle \textcolor{red}{\delta_0} b$};
\draw (293.7,59.34) node [anchor=north west][inner sep=0.75pt]  [font=\scriptsize] [align=left] {$\displaystyle X_{out}$};
\draw (249.94,110.75) node [anchor=north west][inner sep=0.75pt]  [font=\scriptsize] [align=left] {\begin{minipage}[lt]{26.123356pt}\setlength\topsep{0pt}
\begin{center}
Output \\layer
\end{center}
\end{minipage}};
\draw (144.45,77) node [anchor=north west][inner sep=0.75pt]   [align=left] {$\displaystyle \dotsc $};
\end{tikzpicture}
\caption{The architechture of the LSVT network}
\label{fig:architechture}
\end{subfigure}
\caption{Deep neural network for the Learned SVT. The variables in red are learnable parameters.}
\label{fig:DNN}
\end{figure*}

    \section{SVT algorithm} \label{sec:svt}

SVT algorithm solves affine rank minimization formulated in \eqref{eqn:arm} by minimizing the nuclear norm of the matrix which is a surrogate to the rank function. Specifically, the convex optimization problem that the SVT solves is given as \cite{svt}
\begin{subequations}
    \begin{alignat}{2}
    &\!\min_{X \in \mathbb{R}^{d \times d}}        &\qquad& \tau \norm{X}_{tr} + \frac{1}{2} \norm{X}_F^2   \\
    &\text{such that} &      & \mathcal{A}(X)=\bm b
    \end{alignat}  \label{eqn:svtproblem}
\end{subequations} where $\tau>0$ is a constant and the map $\mathcal{A}$ is as defined in \eqref{eqn:lmap}. In theorem 3.1 of \cite{svt}, authors proved that in the limit $\tau$ tending to infinity, the solutions to \eqref{eqn:svtproblem} converge to the matrix with minimum trace norm that is also consistent with the measurements i.e., ($\mathcal{A}(X) = \bm b$). In \cite{arm} authors proved that minimizing the nuclear norm yields the minimum rank solution with high probability (see theorem 3.3 and theorem 4.2 of \cite{arm}). Hence SVT minimizes rank formulated in \eqref{eqn:arm} by solving \eqref{eqn:svtproblem}. 

Note that the problem \eqref{eqn:svtproblem} is a contrained convex optimization problem and the strong duality holds for the problem \cite{boyd}. The Lagrangian function for the problem is given as
\begin{equation}
    \mathcal{L}(X, \bm y) = \tau \norm{X}_{tr} + \frac{1}{2} \norm{X}_F^2 + \bm y^T (\bm b - \mathcal{A}(X))
\end{equation}
where $X$ is the optimization variable and $\bm y$ is the Lagrangian variable. Since strong duality holds for the problem, finding the saddle point $(X^*, \bm y^*)$ of the Lagrangian $\mathcal{L}(X, \bm y)$ gives the solution ($X^*$) to \eqref{eqn:svtproblem}. The saddle point is written as
\begin{equation}
    \underset{\bm y}{\sup} \, \underset{X}{\inf} \, \mathcal{L}(X, \bm y) = \mathcal{L}(X^*,\bm y^*) = \underset{X}{\inf} \, \underset{\bm y}{\sup} \, \mathcal{L}(X, \bm y)
\end{equation}

Authors in \cite{svt} used Uzawa's iterations \cite{uzawa} to find the saddle point of the Lagrangian. Uzawa's iterations starts with an initial $\bm y^0$ and repeats two steps until convergence. In the first step, the minimizer (say $X^k$) of $\mathcal{L}(., \bm y)$ for the given $\bm y$ is found. In the second step, a gradient ascent step is taken along the direction $\bm y$ for the given $X^k$ found in the previous step. These steps are given as follows
\begin{align}
    \begin{cases}
    X^k &= \underset{X}{\arg \min} \, \mathcal{L}(X, \bm y^{k-1}) \\
    \bm y^k &= \bm y^{k-1} + \delta_k \nabla_{\bm y} \mathcal{L}(X^k , \bm y)
    \end{cases} \label{eqn:uzawaiterations}
\end{align} where $\delta_k > 0$ is the step size used for gradient ascent in the $k^{th}$ iteration. The gradient $\nabla_{\bm y} \mathcal{L}(X^k , \bm y)$ is given as $\mathcal{A}(X^k) - \bm b$. The closed form solution to the minimization problem in \eqref{eqn:uzawaiterations} is given as
\begin{equation}
    \underset{X}{\arg \min} \, \mathcal{L}(X, \bm y^{k-1}) = \mathcal{D}_{\tau}( \mathcal{A}^*(\bm y^{k-1})) \label{eqn:minalongX}
\end{equation}
where the operator $\mathcal{A}^*:\mathbb{R}^m \to \mathbb{R}^{d \times d}$ is the adjoint of the linear map $\mathcal{A}$ and is defined as
\begin{equation}
    \mathcal{A}^*(\bm y) = \sum_{i=1}^m y_i A_i^T \label{eqn:adjoint}
\end{equation} and the operator $\mathcal{D}_\tau:\mathbb{R}^{d \times d} \to \mathbb{R}^{d \times d}$ is the singular value thresholding operator and is defined as
\begin{equation}
    \mathcal{D}_{\tau}(X) = U \mathcal{D}_{\tau}(\Sigma)V^T \label{eqn:dtau}
\end{equation}
where the singular value decomposition of $X$ is given as $X = U \Sigma V^T$, $\Sigma$ is the diagonal matrix with the singular values of $X$ in its diagonal positions which we write as $\Sigma = \text{diag}(\sigma_1(X), \dots \sigma_r(X))$ where $r$ represents the rank of $X$. $\mathcal{D}_{\tau}(\Sigma)$ is a diagonal matrix whose non zero entries are found by soft thresolding the entries of $\Sigma$ and is given as $\mathcal{D}_{\tau}(\Sigma) = \text{diag}(\max(\sigma_1(X) - \tau, 0), \dots \max(\sigma_r(X) - \tau, 0))$. 
Using the gradient of $\mathcal{L}$ along $\bm y$ and the equations \eqref{eqn:minalongX}, \eqref{eqn:adjoint} and \eqref{eqn:dtau} Uzawa's iterations \eqref{eqn:uzawaiterations} can be rewritten as
\begin{align}
    \begin{cases}
    X^k &= \mathcal{D}_{\tau}( \mathcal{A}^*(\bm y^{k-1}))\\
    \bm y^k &= \bm y^{k-1} + \delta_k 
    (\bm b - \mathcal{A}(X^k))
    \end{cases} \label{eqn:uzawaiterationsfinal}
\end{align}
where $\delta_k > 0$ is the stepsize to do gradient ascent, $\tau > 0$ is the threshold value used in singular value thresholding operator and the maps $\mathcal{A}$ and $\mathcal{A}^*$ are defined using the measurement matrices $A_1$ through $A_m$ (given in \eqref{eqn:lmap} and \eqref{eqn:adjoint}). 

SVT algorithm repeatedly performs the steps in \eqref{eqn:uzawaiterationsfinal} which requires the knowledge of the right choice of $\delta_k$ and $\tau$ to solve the affine rank minimization problem \eqref{eqn:arm}. In the subsequent sections, we design a trainable deep network based on the steps in \eqref{eqn:uzawaiterationsfinal} for performing affine rank minimization that demands no such tunable parameters. 

    \vspace{-3mm}
\section{Learned SVT} \label{sec:LSVT}

In this section, we design a deep neural network based on the SVT algorithm discussed in the previous section and present a training method to train the network for performing affine rank minimization.

\subsection{Network Architecture} \label{sec:architecture}

Recall that in each iteration of SVT \eqref{eqn:uzawaiterationsfinal}, two steps are performed. We first design a single hidden layer of our network which performs these two steps. To do so, we see that the steps \eqref{eqn:uzawaiterationsfinal} can be rewritten as
\begin{equation}
        \bm y_t = \bm y_{t-1} + \delta_t \left[\bm b - \mathcal{A}(\mathcal{D}_\tau(\mathcal{A}^*(\bm y_{t-1}))) \right] \label{eqn:svtunrolledeqn}
\end{equation} where $\bm y_t$ would represent the output of the $t^{th}$ hidden layer. It can be seen from this reformulation that performing \eqref{eqn:svtunrolledeqn} $T$ times is equivalent to running the SVT algorithm for $T$ iterations. We design the deep network wherein each hidden layer perform \eqref{eqn:svtunrolledeqn} \footnote{$\mathcal{D}_{\tau}(.)$ in \eqref{eqn:svtunrolledeqn} is the singular value thresholding (SVT) operator. Note that the recent work on RPCA \cite{RPCA} also designed a neural network which uses SVT in its layers and backpropagates through it}. To obtain a matrix as the output, we add an output layer at the end of the network which performs the first step of \eqref{eqn:uzawaiterationsfinal}. 

To unroll a fixed number (say $T$) of SVT iterations, we build $T-1$ hidden layers and an output layer and connect them sequentially. Hence, a $T$ layered unrolled network and the SVT algorithm that runs for a fixed $T$ number of iterations are comparable and they are one and the same when the maps $\mathcal{A}, \mathcal{A}^*$ and the parameters $\delta_t$, $\tau_t$ used in both algorithms are same.

We denote by $W_t$ the measurements matrices $A_1$ through $A_m$ used in the maps $\mathcal{A}$ and $\mathcal{A}^*$ and make it learnable. Here $W_t$ is an $m \times d^2$ matrix whose $i^{th}$ row is formed from the entries of $A_i$ such that the measurement vector $\bm b$ given in \eqref{eqn:meas} can be written as $W_t vec(X)$. By making this pair $\{\mathcal{A},\mathcal{A}^*\}$ learnable we try to leverage the power of deep learning to obtain an unrolled variant of SVT which performs better than SVT (which has fixed known pair of $\{\mathcal{A}, \mathcal{A}^*\}$. We denote by $\delta_t$ the step size used in the $t^{th}$ layer and by $\tau_t$ the threshold used in the $t^{th}$ layer and we also make these learnable. With these learnable parameters, the complete network architecture is depicted in Fig. \ref{fig:DNN} where Fig. \ref{fig:hiddenlayers} depicts the hidden layers of our unrolled network and Fig. \ref{fig:outputlayer} depicts the output layer of our network.

\subsection{Training the network} \label{sec:training}

Consider a Learned SVT network with $T$ layers as discussed in previous subsection. The input to this network is $\bm y_0 \in \mathbb{R}^m$ and the output of the network is a matrix $\hat X \in \mathbb{R}^{d \times d}$. From \eqref{eqn:svtunrolledeqn} it can be seen that if $\bm y_{0}$ was a zero vector, $\bm y_1$ would then be $\delta \bm b$. Hence we feed the network with $\bm y_0 = \delta_0 \bm b$ to recover the corresponding matrix $X$. Note that we can also feed the network with zero vector and compare its performance with SVT whose input is also zero vector. We denote by $\Theta$ the set of all learnable parameters in the network i.e., $\Theta = \{ W_1, \dots W_{T}, \delta_0, \delta_1, \dots \delta_{T}, \tau_1, \dots \tau_T \}$. With this notation, the output of the network for a given measurement vector (say $\bm b$) is written as
\begin{equation}
    \hat X = f_{\Theta}(\bm b)
\end{equation}

We denote the training dataset by $\{X^{(i)}, \bm b^{(i)}\}_{i=1}^M$ where $\bm b^{(i)} \in \mathbb{R}^m$ is the measurement vector of the matrix $X^{(i)} \in \mathbb{R}^{d \times d}$ obtained using the known measurement matrices $A_1$ through $A_m$ as described in \eqref{eqn:meas}. For our numerical simulations, which we discuss in the next section, we generate the measurement matrices and $\{X^{(i)}\}$ synthetically. Then we obtain $\{\bm b^{(i)}\}$ using \eqref{eqn:lmap} as $\bm b^{(i)} = \mathcal{A}(X^{(i)} ; A_1, \dots A_m)$. We train our network to minimize the mean squared error (MSE) between the matrices $\{X^{(i)}\}$ in the training dataset and the estimated matrices $f_{\Theta}(\bm b^{(i)})$. The MSE loss is given as
\begin{equation}
    \ell \{ \Theta;\{X^{(i)}, \bm b^{(i)}\} \} = \frac{1}{M} \sum_{i=1}^M \norm{ X^{(i)} - f_{\Theta}(\bm b^{(i)}) }_F^2 \label{eqn:mse}
\end{equation}

\begin{table*}[ht!]
\centering
\resizebox{13cm}{!}{
\begin{tabular}{|c|c|c|c|c|c|c|c|c|c|c|}
\hline
\textbf{\begin{tabular}[c]{@{}c@{}}Iterations/\\ Layers\end{tabular}} & \multicolumn{2}{c|}{\textbf{2}} & \multicolumn{2}{c|}{\textbf{3}} & \multicolumn{2}{c|}{\textbf{4}} & \multicolumn{2}{c|}{\textbf{5}} & \multicolumn{2}{c|}{\textbf{6}} \\ \hline
\textbf{Rank(r)}                                                            & \textbf{SVT}  & \textbf{LSVT}   & \textbf{SVT}  & \textbf{LSVT}   & \textbf{SVT}  & \textbf{LSVT}   & \textbf{SVT}  & \textbf{LSVT}   & \textbf{SVT}  & \textbf{LSVT}   \\ \hline
1                                                                           & 2.3166        & \textbf{0.2234} & 0.8655        & \textbf{0.0687} & 0.4269        & \textbf{0.0300} & 0.2240        & \textbf{0.0174} & 0.1460        & \textbf{0.0158} \\ \hline
2                                                                           & 4.0590        & \textbf{0.1594} & 1.2707        & \textbf{0.0268} & 0.4320        & \textbf{0.0086} & 0.1826        & \textbf{0.0053} & 0.0969        & \textbf{0.0013} \\ \hline
3                                                                           & 5.3233        & \textbf{0.4035} & 1.8576        & \textbf{0.0913} & 0.8466        & \textbf{0.0370} & 0.4532        & \textbf{0.0184} & 0.2748        & \textbf{0.0107} \\ \hline
\end{tabular}
}
\caption{MSE in estimating 10 x 10 matrix by SVT and LSVT for different layers.}
\label{tab:mse10}
\end{table*}

\begin{table*}[ht!]
\centering
\resizebox{13cm}{!}{
\begin{tabular}{|c|c|c|c|c|c|c|c|c|c|c|}
\hline
\textbf{\begin{tabular}[c]{@{}c@{}}Iterations/\\ Layers\end{tabular}} & \multicolumn{2}{c|}{\textbf{2}} & \multicolumn{2}{c|}{\textbf{3}} & \multicolumn{2}{c|}{\textbf{4}} & \multicolumn{2}{c|}{\textbf{5}} & \multicolumn{2}{c|}{\textbf{6}} \\ \hline
\textbf{Rank(r)}                                                            & \textbf{SVT}  & \textbf{LSVT}   & \textbf{SVT}  & \textbf{LSVT}   & \textbf{SVT}  & \textbf{LSVT}   & \textbf{SVT}  & \textbf{LSVT}   & \textbf{SVT}  & \textbf{LSVT}   \\ \hline
2                                                                           & 4.6028        & \textbf{0.5565} & 1.4991        & \textbf{0.1630} & 0.7064        & \textbf{0.0693} & 0.3738        & $\mathbf{0.2690}^@$ & 0.2538        & $\mathbf{0.1364}^@$ \\ \hline
4                                                                           & 8.3921        & \textbf{0.4007} & 2.5227        & \textbf{0.0784} & 0.8701        & \textbf{0.0893} & 0.3832        & \textbf{0.0942} & 0.2224        & \textbf{0.0097} \\ \hline
6                                                                           & 10.9441       & \textbf{1.008}  & 3.9163        & \textbf{0.2498} & 1.7994        & \textbf{0.2904} & 0.9914        & \textbf{0.0471} & 0.6328        & \textbf{0.0564} \\ \hline
\end{tabular}
}
\caption{MSE in estimating 20 x 20 matrix by SVT and LSVT for different layers.}
\label{tab:mse20}
\end{table*}

\subsection{Initialization and training}

We initialize the network's trainable parameters $\Theta$ with those used in SVT algorithm. We initialize each of $\{W_t\}$ with the measurement matrices $A_1$ through $A_m$. Each of the stepsizes $\{\delta_0, \delta_1, \dots \delta_T\}$ is initialized to $1.2 \times \frac{d^2}{m}$ and each of the thresholds $\{\tau_1, \dots \tau_T\}$ is initializes to $5 \times d$ as these values were adapted by the authors of SVT for better results. Since we initialize our network with the parameters used in SVT algorithm, our networks performs similar to SVT (with fixed $T$ number of iterations) initially. We also initialize the parameters $\{\delta_0, \delta_1, \dots \delta_T \dots \tau_1 \dots \tau_T\}$ with values other than the ones used by authors of SVT, to see the dependence of SVT and LSVT on these parameters. The initial error that the network incurs is same as that incurred by SVT. We use gradient based optimizer ADAM \cite{adam} to minimize the loss \eqref{eqn:mse} and train the network.
    \vspace{-3mm}
\section{Numerical simulation} \label{sec:simulations}
We design and train the proposed Learned SVT network to estimate $d \times d$ matrices of rank $r$ from its $m$ linear measurements. The linear map $\mathcal{A}$ that defines the linear measurements is synthetically generated and is fixed through out the experiments. To do this we randomly generate the matrices $A_1$ through $A_m$ such that $\tr[A_i^T A_j] = \delta_{ij}$. We generate 61,000 ground truth data $\{X^{(i)} \in \mathbb{R}^{d \times d}, \bm b^{(i)} \in \mathbb{R}^m \}_{i=1}^{61,000}$ where $\{X^{(i)}\}$ are of rank-r and $\{\bm b^{(i)}\}$ are the corresponding measurement vectors. To do this, we generate $P^{(i)} \in \mathbb{R}^{d \times r}$ and $Q^{(i)} \in \mathbb{R}^{r \times d}$ randomly with each entry of $P^{(i)}$ and $Q^{(i)}$ generated from $\mathcal{N}(0,2)$ then we get $X^{(i)}$ by multiplying $P^{(i)}$ and $Q^{(i)}$. $\bm b^{(i)}$ are obtained by measuring $X^{(i)}$ using the measurement map generated. We use PyTorch to design and train our LSVT network. PyTorch automatically computes the gradients of the loss function with respect to the network parameters using computational graphs and Autograd functionality. We use stochastic gradient descent based ADAM optimizer with a learning rate of $10^{-4}$ to train the network. We perform mini batch training with a minibatch size of 1000.  Of the 61,000 ground truth data, we use 50,000 to train the network, 10,000 to validate the network. We validate the network every time the network parameter gets updated and we stop the training process when this validation loss doesn't decrease over the course of training. Once the training is over, we freeze the network parameters and test the network with the remaining 1000 ground truth data and the resulted MSE are reported.

LSVT is designed and trained to estimate $10 \times 10$ matrices of rank 1, 2 and 3 with different layers. We sampled the rank-1 matrix with a oversampling ratio of $3$. For rank-2 and rank-3 matrix, 90 linear measurements were obtained. The MSE in estimating the matrices by both SVT and LSVT algorithms are compared in Table \ref{tab:mse10}. We also estimated $20 \times 20$ matrices of ranks 2, 4 and 6. Rank-2 matrices were sampled with oversampling ratio 3. 350 linear measurements were obtained for rank-4 and rank-6 matrices. The corresponding MSE values are compared in Table \ref{tab:mse20}. The learning rate of $10^{-4}$ is used to train all the networks except for the values super-scripted by $@$ where $10^{-5}$ was used. All MSE values are reported by averaging over 1000 instances. It can be seen from Tables \ref{tab:mse10} and \ref{tab:mse20} that the Learned LSVT performs much better in terms of MSE in estimating the matrices compared with the original SVT algorithm.

To study the dependence of SVT and LSVT on the threshold value ($\tau$) and stepsize $(\delta)$ both algorithms are simulated with different thresholds ($\tau$) and stepsizes ($\delta$) to estimate $20 \times 20$ matrices of rank-2 with a oversampling ratio of 3. SVT uses these parameters and performs 4 SVT iterations, while 4-layered LSVT is initialized with these parameters and trained. The corresponding MSE values are tabulated in Table \ref{tab:svtlsvtmultiparas} where the $*$ denoted values are the ones used by the authors of SVT. From Table \ref{tab:svtlsvtmultiparas} it can be seen that even with different initial parameters ($\tau ,\delta$) LSVT always performs better than SVT in terms of MSE.
\begin{table}[h!]
\centering
\resizebox{6cm}{!}{
\begin{tabular}{|c|c|c|}
\hline
\textbf{Parameters}             & \textbf{SVT} & \textbf{LSVT}   \\ \hline
$\tau = 5, \delta = 1$          & 2.2477       & \textbf{0.1831}   \\ \hline
$\tau = 50, \delta = 0.5$       & 6.4043       & \textbf{0.0479} \\ \hline
$\tau = 50, \delta = 2.10^*$    & 0.2536       & \textbf{0.0483} \\ \hline
$\tau = 100^*, \delta = 2.10^*$ & 0.7064       & \textbf{0.0693} \\ \hline
$\tau = 200, \delta = 5$        & 7.5677       & \textbf{0.4287} \\ \hline
$\tau = 300, \delta = 5$        & 8.1711       & \textbf{1.2766} \\ \hline
\end{tabular}
}
\caption{Comparing MSE in estimating matrices by SVT and LSVT when different parameters are used.}
\label{tab:svtlsvtmultiparas}
\end{table}

\vspace{-5mm}
\section{Conclusion} \label{sec:conclusion}
We designed a trainable deep neural network called LSVT by appropriately unrolling the SVT algorithm. The proposed LSVT with fixed $T$ layers estimates matrices with significantly lesser mean squared error (MSE) compared with MSE incurred by SVT with fixed $T$ iterations. We also showed that LSVT outperforms SVT even when both algorithms use different values for the step sizes and thresholds rather than those suggested by the authors of SVT \cite{svt}. 
	\bibliographystyle{IEEEtran}
	\bibliography{references.bib}

\begin{thebibliography}{10}
\providecommand{\url}[1]{#1}
\csname url@samestyle\endcsname
\providecommand{\newblock}{\relax}
\providecommand{\bibinfo}[2]{#2}
\providecommand{\BIBentrySTDinterwordspacing}{\spaceskip=0pt\relax}
\providecommand{\BIBentryALTinterwordstretchfactor}{4}
\providecommand{\BIBentryALTinterwordspacing}{\spaceskip=\fontdimen2\font plus
\BIBentryALTinterwordstretchfactor\fontdimen3\font minus
  \fontdimen4\font\relax}
\providecommand{\BIBforeignlanguage}[2]{{%
\expandafter\ifx\csname l@#1\endcsname\relax
\typeout{** WARNING: IEEEtran.bst: No hyphenation pattern has been}%
\typeout{** loaded for the language `#1'. Using the pattern for}%
\typeout{** the default language instead.}%
\else
\language=\csname l@#1\endcsname
\fi
#2}}
\providecommand{\BIBdecl}{\relax}
\BIBdecl

\bibitem{localizationedm}
H.~{Zhang}, Y.~{Liu}, and H.~{Lei}, ``Localization from incomplete euclidean
  distance matrix: Performance analysis for the svd–mds approach,''
  \emph{IEEE Transactions on Signal Processing}, vol.~67, no.~8, pp.
  2196--2209, 2019.

\bibitem{environmental}
J.~{He}, Y.~{Zhou}, and G.~{Sun}, ``Environmental monitoring in wireless sensor
  networks using structured matrix completion,'' in \emph{GLOBECOM 2020 - 2020
  IEEE Global Communications Conference}, 2020, pp. 1--5.

\bibitem{lowcostwsnmc}
K.~{Xie}, L.~{Wang}, X.~{Wang}, G.~{Xie}, and J.~{Wen}, ``Low cost and high
  accuracy data gathering in wsns with matrix completion,'' \emph{IEEE
  Transactions on Mobile Computing}, vol.~17, no.~7, pp. 1595--1608, 2018.

\bibitem{arraysignal}
Z.~{Weng} and X.~{Wang}, ``Low-rank matrix completion for array signal
  processing,'' in \emph{2012 IEEE International Conference on Acoustics,
  Speech and Signal Processing (ICASSP)}, 2012, pp. 2697--2700.

\bibitem{beamforming}
J.~{Wen}, X.~{Zhou}, B.~{Liao}, C.~{Guo}, and S.~{Chan}, ``Adaptive beamforming
  in an impulsive noise environment using matrix completion,'' \emph{IEEE
  Communications Letters}, vol.~22, no.~4, pp. 768--771, 2018.

\bibitem{mimomc}
E.~{Vlachos}, G.~C. {Alexandropoulos}, and J.~{Thompson}, ``Massive mimo
  channel estimation for millimeter wave systems via matrix completion,''
  \emph{IEEE Signal Processing Letters}, vol.~25, no.~11, pp. 1675--1679, 2018.

\bibitem{exact}
E.~J. Cand{\`e}s and B.~Recht, ``Exact matrix completion via convex
  optimization,'' \emph{Foundations of Computational mathematics}, vol.~9,
  no.~6, pp. 717--772, 2009.

\bibitem{gross}
D.~Gross, Y.-K. Liu, S.~T. Flammia, S.~Becker, and J.~Eisert, ``Quantum state
  tomography via compressed sensing,'' \emph{Physical review letters}, vol.
  105, no.~15, p. 150401, 2010.

\bibitem{sevenqubit}
C.~Riofr{\'\i}o, D.~Gross, S.~T. Flammia, T.~Monz, D.~Nigg, R.~Blatt, and
  J.~Eisert, ``Experimental quantum compressed sensing for a seven-qubit
  system,'' \emph{Nature communications}, vol.~8, no.~1, pp. 1--8, 2017.

\bibitem{arm}
B.~Recht, M.~Fazel, and P.~A. Parrilo, ``Guaranteed minimum-rank solutions of
  linear matrix equations via nuclear norm minimization,'' \emph{SIAM review},
  vol.~52, no.~3, pp. 471--501, 2010.

\bibitem{lowrankanybasisgross}
D.~{Gross}, ``Recovering low-rank matrices from few coefficients in any
  basis,'' \emph{IEEE Transactions on Information Theory}, vol.~57, no.~3, pp.
  1548--1566, 2011.

\bibitem{powerfactorization}
J.~P. {Haldar} and D.~{Hernando}, ``Rank-constrained solutions to linear matrix
  equations using powerfactorization,'' \emph{IEEE Signal Processing Letters},
  vol.~16, no.~7, pp. 584--587, 2009.

\bibitem{als}
D.~{Zachariah}, M.~{Sundin}, M.~{Jansson}, and S.~{Chatterjee}, ``Alternating
  least-squares for low-rank matrix reconstruction,'' \emph{IEEE Signal
  Processing Letters}, vol.~19, no.~4, pp. 231--234, 2012.

\bibitem{saresepowerfactorization}
K.~{Lee}, Y.~{Wu}, and Y.~{Bresler}, ``Near-optimal compressed sensing of a
  class of sparse low-rank matrices via sparse power factorization,''
  \emph{IEEE Transactions on Information Theory}, vol.~64, no.~3, pp.
  1666--1698, 2018.

\bibitem{rankonemeas}
Y.~Li, Y.~Sun, and Y.~Chi, ``Low-rank positive semidefinite matrix recovery
  from corrupted rank-one measurements,'' \emph{IEEE Transactions on Signal
  Processing}, vol.~65, no.~2, pp. 397--408, 2016.

\bibitem{psdrlm}
Q.~Zheng and J.~Lafferty, ``A convergent gradient descent algorithm for rank
  minimization and semidefinite programming from random linear measurements,''
  in \emph{Advances in Neural Information Processing Systems}, C.~Cortes,
  N.~Lawrence, D.~Lee, M.~Sugiyama, and R.~Garnett, Eds., vol.~28.\hskip 1em
  plus 0.5em minus 0.4em\relax Curran Associates, Inc., 2015.

\bibitem{svt}
J.-F. Cai, E.~J. Cand{\`e}s, and Z.~Shen, ``A singular value thresholding
  algorithm for matrix completion,'' \emph{SIAM Journal on optimization},
  vol.~20, no.~4, pp. 1956--1982, 2010.

\bibitem{unrollingsurvey}
V.~Monga, Y.~Li, and Y.~C. Eldar, ``Algorithm unrolling: Interpretable,
  efficient deep learning for signal and image processing,'' \emph{IEEE Signal
  Processing Magazine}, vol.~38, no.~2, pp. 18--44, 2021.

\bibitem{boyd}
S.~Boyd, S.~P. Boyd, and L.~Vandenberghe, \emph{Convex optimization}.\hskip 1em
  plus 0.5em minus 0.4em\relax Cambridge university press, 2004.

\bibitem{uzawa}
K.~Arrow, H.~Chenery, H.~Azawa, K.~M.~R. Collection, L.~Hurwicz, H.~Uzawa,
  S.~Johnson, S.~Karlin, T.~Marschak, and R.~Solow, \emph{Studies in Linear and
  Non-linear Programming}, ser. Stanford mathematical studies in the social
  sciences.\hskip 1em plus 0.5em minus 0.4em\relax Stanford University Press,
  1958.

\bibitem{RPCA}
O.~{Solomon}, R.~{Cohen}, Y.~{Zhang}, Y.~{Yang}, Q.~{He}, J.~{Luo}, R.~J.~G.
  {van Sloun}, and Y.~C. {Eldar}, ``Deep unfolded robust pca with application
  to clutter suppression in ultrasound,'' \emph{IEEE Transactions on Medical
  Imaging}, vol.~39, no.~4, pp. 1051--1063, 2020.

\bibitem{adam}
D.~P. Kingma and J.~Ba, ``Adam: {A} method for stochastic optimization,'' in
  \emph{3rd International Conference on Learning Representations, {ICLR} 2015,
  San Diego, CA, USA, May 7-9, 2015, Conference Track Proceedings}, Y.~Bengio
  and Y.~LeCun, Eds., 2015.

\end{thebibliography}
\end{document}